\documentclass[10pt,twocolumn,letterpaper]{article}

\usepackage{cvpr}
\usepackage{times}
\usepackage{epsfig}
\usepackage{graphicx}
\usepackage{amsmath}
\usepackage{amssymb}
\usepackage{bbm}
\usepackage{color}
\usepackage{esvect}
\usepackage{enumitem}
\usepackage{stfloats}

\usepackage[breaklinks=true,bookmarks=false]{hyperref}

\cvprfinalcopy 

\def\cvprPaperID{6191} 

\newcommand{\bd}{\mathbf{d}}
\newcommand{\bg}{\mathbf{g}}
\newcommand{\bG}{\mathbf{G}}

\newcommand{\bT}{\mathbf{T}}

\renewcommand{\vv}[1]{\mathbf #1}

\newcommand{\IDF}{{\bf IDF}}
\newcommand{\MOTA}{{\bf MOTA}}

\newcommand{\ours}{{\bf OURS}}

\newcommand{\rnn}{{\bf RNN}}
\newcommand{\sort}{{\bf SORT}}
\newcommand{\ptrack}{{\bf PTRACK}}
\newcommand{\lp}{{\bf LP2D}}
\newcommand{\reid}{{\bf REID}}
\newcommand{\cdsc}{{\bf CDSC}}
\newcommand{\mht}{{\bf MHT}}
\newcommand{\bipcc}{{\bf BIPCC}}

\newcommand{\dman}[0]{{\bf DMAN}}
\newcommand{\jcc}[0]{{\bf JCC}}
\newcommand{\motdt}[0]{{\bf MOTDT17}}
\newcommand{\mhtblstm}[0]{{\bf MHTBLSTM}}
\newcommand{\edmt}[0]{{\bf EDMT17}}
\newcommand{\fwt}[0]{{\bf FWT}}

\newif\ifdraft
\draftfalse

\ifdraft

\newcommand{\PF}[1]{{\color{red}{pf: \bf #1}}}

\newcommand{\FF}[1]{{\color{red}{ff: \bf #1}}}
\newcommand{\MS}[1]{{\color{blue}{ms: \bf #1}}}

\newcommand{\AM}[1]{{\color{green}{am: \bf #1}}}
\else
 \newcommand{\PF}[1]{}
 
 \newcommand{\FF}[1]{}
 
 \newcommand{\MS}[1]{}
 \newcommand{\AM}[1]{}
 
\fi

\newcommand{\comment}[1]{}
\newcommand{\parag}[1]{\vspace{-3mm}\paragraph{#1}}

\newcommand{\Duke}[0]{{\bf DukeMTMC}}

\newcommand{\MotFive}[0]{{\bf MOT15}}
\newcommand{\MotSeven}[0]{{\bf MOT17}}
\newcommand{\PathTrack}[0]{{\bf PathTrack}}

\begin{document}

\title{Eliminating Exposure Bias and  Loss-Evaluation Mismatch in Multiple Object Tracking}

\author{Andrii Maksai \qquad Pascal Fua\\
Computer Vision Laboratory, \'Ecole Polytechnique F\'ed\'erale de Lausanne (EPFL)\\
{\tt\small \{andrii.maksai, pascal.fua\}@epfl.ch}
}

\maketitle

\begin{abstract}

  Identity  Switching remains  one  of the  main  difficulties Multiple  Object
Tracking (MOT) algorithms  have to deal with.  Many state-of-the-art approaches
now  use sequence  models  to solve  this  problem but  their  training can  be
affected by biases that decrease their  efficiency. In this paper, we introduce
a new training procedure that confronts the algorithm to its own mistakes while
explicitly  attempting to  minimize the  number of  switches, which  results in
better training.

We propose an iterative scheme of building  a rich training set and using it to
learn a  scoring function  that is  an explicit proxy  for the  target tracking
metric.  Whether using  only simple  geometric features  or more  sophisticated
ones  that also  take appearance  into  account, our  approach outperforms  the
state-of-the-art on several MOT benchmarks.

\end{abstract}

\vspace{-0.4cm}

\section{Introduction}
\label{sec:intro}
\vspace{-2mm}


\begin{figure*}[!b]
  \includegraphics[width=\textwidth,trim={0 9.7cm 0cm 2.7cm},clip]{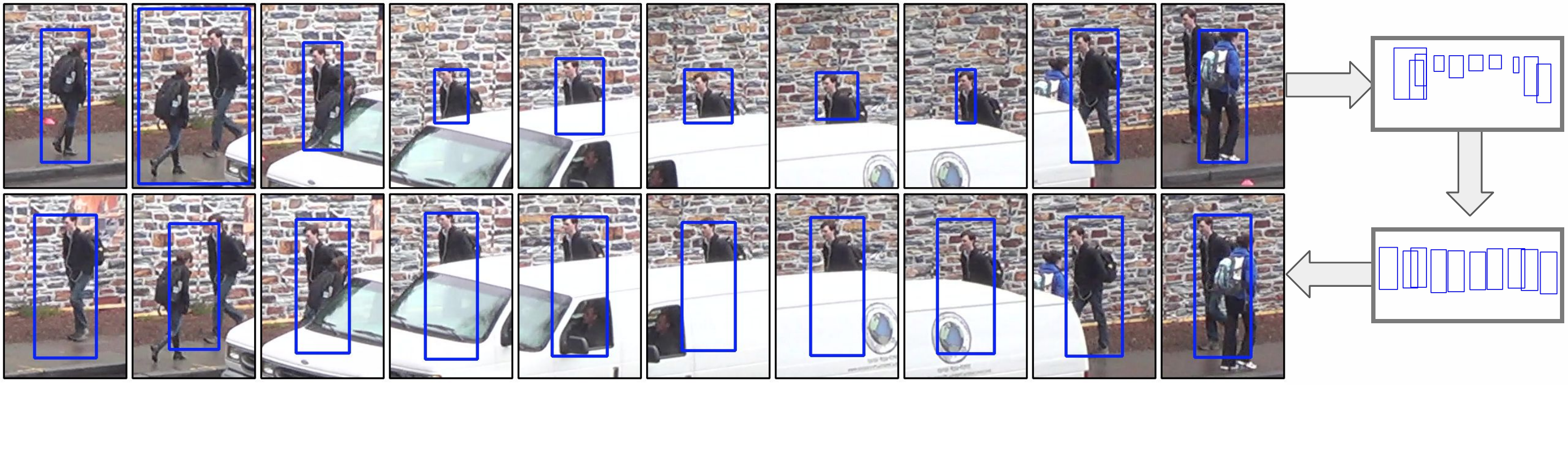}
  \vspace{-7mm}
  \caption{\small Keeping track in a difficult situation. {\bf Top row:} Because of the occlusion created by the passing car, a tracker can easily return a trajectory that includes several identity switches. The corresponding bounding boxes inside camera's field of view are shown on the right. {\bf Bottom row:} Our algorithm not only eliminates identity switches but also regresses to a set of much tighter bounding boxes. In this example our algorithm did it solely on the basis of simple geometric features without requiring the use of appearance information. }
  \label{fig:teaser}
\end{figure*}

A common concern  in Multi Object Tracking (MOT)  approaches is to prevent
identity  switching, the  erroneous  merging of  trajectories corresponding  to
different  targets into  a single  one. This  is difficult  in crowded  scenes,
especially  when  the  appearances  of  the individual  target  objects  are  not
distinctive enough. Many recent approaches rely on tracklets---short trajectory
segments---rather  than individual  detections,  to keep  track  of the  target
objects. Tracklets can  be merged into longer trajectories, which  can be split
again when  an identity  switch occurs.

Most    state-of-the-art   approaches    rely   on    deep   networks,    often
in    the    form    of    RNN   architectures    that    operate    on    such
tracklets~\cite{Long18,Henschel18,Keuper18,Yoon18,Kim18}.     This     requires
training the sequence  models and is subject  to one or both  of two well-known
problems, which our approach overcomes:
\begin{itemize}[leftmargin=*]

  \vspace{-0.1cm}
  \item {\bf Loss-evaluation  mismatch.} It occurs when  training by optimizing
a metric  poorly  aligned  with the  actual  desired performance  during
inference. In MOT, one example is the use of a classification  loss to  create trajectories
optimal  for a  tracking-specific metric,  such as  \MOTA~\cite{Bernardin08} or
\IDF{}~\cite{Ristani16}. To eliminate this mismatch, we introduce an original way to score
tracklets that is an  explicit proxy for the \IDF{} metric  and can be computed
without the ground truth.  We use it to identify how  confidently the person is
tracked, predict tighter bounding box locations,  and estimate whether
the real trajectory extends beyond the observed tracklet.


  \vspace{-0.1cm}
  \item {\bf Exposure bias. } It stems  from the model not being exposed to its
own  errors during  training and  results in  very different  data distribution
observed  during  training  and  inference/tracking. We  remove  this  bias  by
introducing a much  more exhaustive, yet computationally  feasible, approach to
exploiting the data while training the  model. To this end, during training, we
do not limit ourselves to only using tracklets made of detections of one or two
people as in~\cite{Milan17,Ma18,Sadeghian17}. Instead, we consider any grouping
of tracklets  produced by the tracking  algorithm to be a  potential trajectory
but prevent  a combinatorial explosion  by controlling the number  of tracklets
that share many common detections. This yields a much richer training dataset,
solves the exposure bias problem, and enables our algorithm to handle confusing
situations in which  a tracking algorithm may easily switch  from one person to
the next  or miss  someone altogether.  Fig.~\ref{fig:teaser} depicts  one such
case. Note that this can be done  even when appearance information is {\it not}
available.
\vspace{-0.1cm}

\end{itemize}
Our contribution is therefore a solution  to these two problems. By integrating
it into  an algorithm  that only uses  very simple  features----bounding boxes,
detector confidence---we outperform other approaches that do not use appearance
features either. By also exploiting appearance-based features, we similarly outperform
state-of-the-art approaches that do.  Taken together, these results demonstrate
the effectiveness of our training procedure.

In  the remainder  of this  paper,  we first  briefly review  related work  and
current approaches to mitigating loss-evaluation mismatch and exposure bias. We
then  introduce  our approach  to  tracking;  it  is  a variation  of  multiple
hypothesis tracking  designed for  learning to  efficiently score  the tracklets.
Next, we  describe the exact  form of our scoring  function and its  ability to
reduce both mismatch and bias. Finally, we present our results.

\vspace{-0.2cm}

\section{Related work}
\label{sec:related}

Multiple  Object  Tracking  (MOT)  has   a  long  tradition,  going  back  many
years  for   applications  such   as  radar   tracking~\cite{Blackman86}.  With
the  recent   improvements  of  object  detectors,   the  tracking-by-detection
paradigm~\cite{Andriluka08}  has  become a  {\it de facto}  standard  and has  proven
effective  for many  applications  such  as surveillance  or  sports player tracking.  It involves first detecting  the target  objects in
individual  frames,  associating  these  detections  into  short  but  reliable
trajectories  known  as  tracklets,  and  then  concatenating  these  tracklets
into  longer  trajectories.  They  can  then   be  used  to  solve  tasks  such
as  social  scene understanding~\cite{Alahi16,Bagautdinov17},  future  location
prediction~\cite{Lee17a}, or human dynamic modeling~\cite{Fragkiadaki15}.

While  grouping individual  detections  into trajectories  it  is difficult  to
guarantee that  each resulting trajectory represents  a {\it whole} track  of a
{\it single} individual , that is, that there are no identity switches.

Many approaches rely on appearance~\cite{Held16,Leal-Taixe16,Zhai16,Zhang17,Chen17b,Lin17b,Ristani18}, motion~\cite{Dickle13}, or social cues~\cite{Hu08,Pellegrini09}. They are mostly used to associate pairs of detections, and only account for very short-term correlations. However, since people trajectories are often predictable over many frames once a few have been seen, superior performance could be obtained by modeling behavior over longer time periods~\cite{Iqbal17,Koh17, Maksai17}. Increasing availability of annotated training data and  benchmarks, such as  MOT15-17~\cite{Leal-Taixe15,Milan16b}, DukeMTMC~\cite{Ristani16}, PathTrack~\cite{Manen17}, and Wildtrack~\cite{Chavdarova18a} now makes it possible to learn  the data association models required to leverage this knowledge. Since this is what our method does, we briefly review here a few state-of-the-art approaches to achieving this goal.

\subsection{Modeling Longer Sequences}

The  work   of~\cite{Ondruska16a,Ondruska16b}  is  one  of   the  first  recent
approaches to modeling long trajectories  using a recurrent neural network. The
algorithm  estimates  ground-plane occupancy,  but  does  not perform  explicit
data  association.~\cite{Milan17}  presented  an approach  to  performing  data
association  without  using  appearance   features  by  predicting  the  future
location  of   the  target.  Several   MOT  approaches  have   followed,  using
sequence  models  to  make  data   association  more  robust  for  the  purpose
of  people  re-identification~\cite{Sadeghian17,Ma18}, learning  better  social
models~\cite{Alahi16},  forecasting  future  locations~\cite{Lee17a,Xiang18a}  or
joint detection, tracking, and activity recognition~\cite{Bagautdinov17}.

These models are usually trained on  sample trajectories that perfectly match a
single person's  trajectory or only  marginally deviate from that,  making them
vulnerable to exposure bias. Furthermore, the loss function is usually designed
primarily for localization  or identification rather then fidelity  to a ground
truth trajectory. This introduces a  loss-evaluation mismatch with the metric,
usually \IDF~\cite{Ristani16}  or \MOTA~\cite{Bernardin08}, which reflects more
reliably the desirable behavior of the algorithm.

Most   state-of-the-art   approaches  that   use   sequence   models  rely   on
one  of   two  optimization   techniques,  hierarchical  clustering   for  data
association~\cite{Tesfaye17,Zhang17,Ristani16,Long18,Henschel18,Keuper18}    or
multiple hypothesis  tracking~\cite{Yoon18,Kim18,Chen17c}. The  former involves
valid groups of observations without  shared hypotheses while the latter allows
for conflicting  sets of hypotheses to  be present until the  final solution is
found. The approach most similar to our is that of~\cite{Kim18}. It also uses a
combination of  multiple hypothesis tracker  and a sequence model  for scoring.
However, the training  procedure mostly relies on ground  truth information and
is  therefore more  subject to  exposure bias.  Another closely related method is that of~\cite{Milan17} that
trains a sequence model for data association solely from geometric features and
is therefore well-suited for comparison with  our approach when also using only
geometric cues.  These methods are all recent and collectively represent the current state-of-the-art.
In Section~\ref{sec:results}, we  will therefore treat them as
baselines against which we can compare our approach.

\subsection{Reducing  Bias and Loss-Evaluation Mismatch}

Since exposure bias  and loss-evaluation mismatch are also  problems in Natural
Language  Processing  (NLP)~\cite{Vinyals15}  and   in  particular  in  machine
translation~\cite{Wu16}, several methods have been  proposed in these fields to
reduce it~\cite{Ranzato15,Bengio15}.  Most of them, however,  operate under the
assumption that output  sequences can comprise any character  from a predefined
set. As a result, they typically  rely on a beam-search procedure, which itself
frequently uses  a language model to  produce a diverse set  of candidates that
contains the correct one. More generally, techniques that allow training models
without  differentiable relation  between  inputs and  outputs  such as  policy
gradient~\cite{Williams97},  straight-through estimation~\cite{Bengio13b},  and
Gumbel-Softmax~\cite{Jang16} can be seen as methods reducing exposure bias.

Unfortunately, in the case of MOT,  the detections form a spatio-temporal graph
in which  many nearly  identical trajectories  can be  built. This can easily
overwhelm  standard  beam-search  techniques:  when limiting  oneself  to  only
the  top  scoring candidates  to  prevent  a  combinatorial explosion,  it  can
easily  happen that  only  a  set of  very  similar  but spurious  trajectories
will  be considered  and  the real  one  ignored. This  failure  mode has  been
addressed  in  the  context  of  single-object  tracking  and  future  location
prediction  in  ~\cite{Supancic17,Ma17b}  with  a tracking  policy  learned  by
reinforcement  learning and  in~\cite{Chu17} by  introducing a  spatio-temporal
attention mechanism over a batch of images, thus ensuring that within the batch
there is no exposure bias. Instead, the algorithm relies on historical positive
samples  from already  obtained  tracks,  thus re-introducing  it.  For MOT,  a
reinforcement  learning-based  approach  has  been  proposed~\cite{Xiang15}  to
decide  whether  to  create  new  tracklets or  terminate  old  ones.  This  is
also  addressed  in~\cite{Sadeghian17}  but  the learning  of  sequence  models
is  done  independently  and  is  still  subject  to  exposure  bias.  Approach
of~\cite{Maksai17} attempts  to explicitly optimize  for the \IDF{}  metric. It
does so  by refining  the output  of other tracking  methods. This  reduces the
loss-evaluation mismatch  but the sequence  scoring model is  hard-coded rather
than learned and we will show that learning it yields better results.

\vspace{-0.2cm}

\section{Tracklet-Based Tracking}
\label{sec:tracking}

Our approach to tracking relies on creating and merging tracklets to build high-scoring trajectories as in multiple hypothesis tracking~\cite{Kim15}. In this section, we formalize it and describe its components, assuming that the scoring function is given. The scoring function and how it is learned will be discussed in the following section.

\subsection{Formalization}
\label{sec:formalization}

Let us consider a  video sequence made of $N$ frames, on which  we run a people
detection algorithm on each frame individually. This yields a set $\mathcal{D}$
of  people detections  $\bd_i \in  \mathbbm{R}^4$, where  the four  elements of
$\bd_i$ are  the coordinates of  the corresponding  bounding box in  the image.
We  represent  a  tracklet  $\bT$  as  a  $4  \times  N$  matrix  of  the  form
$\left[\bd_1,\bd_2,\ldots,\bd_N\right]$.  In  practice, tracklets  only  rarely
span the whole sequence.  We handle this by setting $\bd_n$  to zero for frames
in  which the  person's location  is unknown.  The first  non-zero column  of a
tracklet is therefore its start and the last its end. Two tracklets $\bT_1$ and
$\bT_2$ can be  merged into a single one  if there is no single  frame in which
they contain different detections.

Let $\Phi: \mathbbm{R}^{4 \times N}  \rightarrow \mathbbm{R}^{F  \times N}$  be a {\it feature} function  that  assigns a  feature vector  of dimension $F$  to each column of  a tracklet. In practice,  these features can be bounding box coordinates, confidence level, and shift from the nearest detection in a previous frame. They can also be image-based features associated to the detection and we list them all in Section~\ref{sec:protocol}. Let us further assume that we can compute from these features a score $S(\Phi(\bT))$ that is  high when the  tracklet truly represents a single person's trajectory and low otherwise. Tracking can  then be understood as building  the  set of  non-overlapping  tracklets  $\bT_j$ that  maximizes  the objective function
\begin{equation}
 \sum_j S(\Phi(\bT_j)) \; .
 \label{eq:objFunc}
\end{equation}
In the remainder of this section,  we will assume that $S$ is given and assigns low scores to the wide range of bad candidate trajectories that can be generated, and high scores to the real ones.

\subsection{Creating and Merging Tracklets}
\label{sec:tracklets}

We iteratively  merge tracklets  to create  ever longer  candidate trajectories
that  include the  real ones  while suppressing  many candidates  to prevent  a
computationally infeasible  combinatorial explosion. We then  select an optimal
subset greedily. We consider two trajectories  to be overlapping when they have
a large  intersection over  union. More  specifically, if  the total  number of
pixels shared by bounding boxes of the two tracklets, normalized by the minimum
of the  sum of areas of  bounding boxes in each  of them, is above  a threshold
$C_{IoU}$. We also  eliminate tracklets that are either shorter  than $N$ - the
length of  the batch, or  whose score  is below another  threshold $C_{score}$.
$C_{IoU}$ and $C_{score}$ are hyper-parameters that we estimate on a validation
set.
Outlined procedure involves the two main steps described below.

\subsubsection{Generating Candidate Trajectories}
\label{sec:candidates}

The set of candidate trajectories must contain the real ones but its size must be kept small enough to prevent a combinatorial explosion. To this end, given the initial set of detections $\mathcal{D}$, which we take to be the initial tracklet set.

We then iterate the following two steps for $n = 2, \dots, N$.
\begin{enumerate}

\vspace{-1mm}
\item {\bf  Growing:} Merge pairs  of tracklets that  can be merged  and result
would be  bigger than the  biggest of  two by 1.  Tracklets with  $k_1$ and
$k_2$ non-zero  detections yields a  tracklet of $\max(k_1,  k_2)+1$ non-zero
detections, that includes non-zero detections from both of them.

  \vspace{-1mm}
\item  {\bf Pruning:}  Given tracklet $\bT_1$, for all  $\bT_2$ that were merged with it during growing phase, only retain the merger that maximizes the score  $S(\Phi(\cdot))$.

\end{enumerate}
This process keeps  the number of hypotheses linear with  respect to the number
of detections. Yet,  it retains a candidate for every  possible detection. This
prevents  the algorithm  from losing  people and  terminating trajectories  too
early  even if  mistakes are  made early  in the  pruning process.  We give  an
example in Fig.~\ref{fig:network}, (b). In  appendix, we compare this heuristic
to several  others and show  that it  is effective at  preventing combinatorial
explosion without losing valid hypotheses.

\subsubsection{Selecting Candidates}

Given the resulting set of tracklets, we want to select a
compatible subset that maximizes our objective  function. To this end we select
a subset  of hypotheses  with the  best possible  sum of  scores, subject  to a
non-overlapping  constraint. We  do this  greedily, starting  with the  highest
scoring trajectories. As discussed in the appendix, we also tried
a  more sophisticated  approach  that casts  it as  an  integer program  solved
optimally, and the results are similar.

\vspace{-0.2cm}

\section{Learning to Score}

The scoring function $S(\Phi(\cdot))$ of Eq.~\ref{eq:objFunc} is a the heart of the tracking procedure of Section~\ref{sec:tracking}. When we create and merge tracklets, we want it to favor those that can be associated to a single person without identity switch, that is, those that score well in terms of the \IDF{} metric. We choose  \IDF{} over other popular alternative such as  \MOTA{} because it has been shown to be more sensitive to identity switches~\cite{Ristani16}.

In the remainder of this section, we first define $S$, which we implement using the deep network depicted by Fig.~\ref{fig:network}(a).  We then describe how we train it.


\subsection{Defining the Scoring Function}
\label{sec:score}


\begin{figure*}
  \begin{tabular}{cc}
    \includegraphics[width=\columnwidth,trim={0 5cm 0 0},clip]{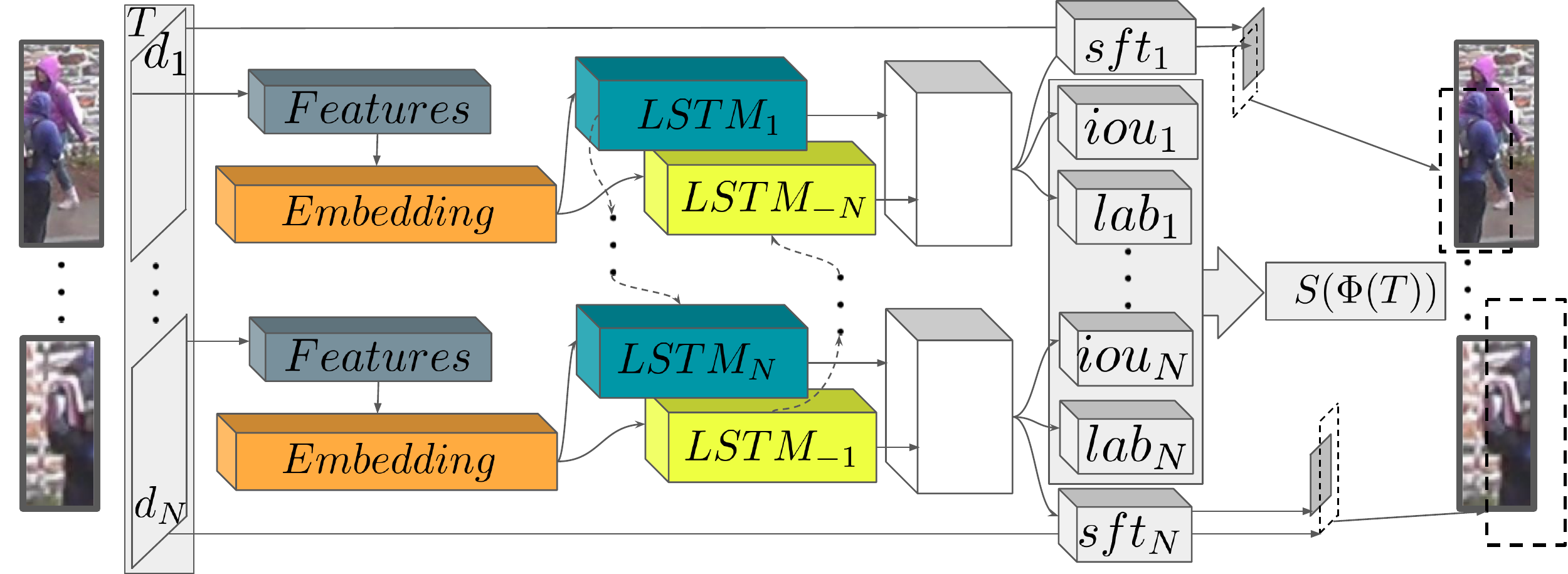} &
    \includegraphics[width=\columnwidth,trim={0 0 0 0},clip]{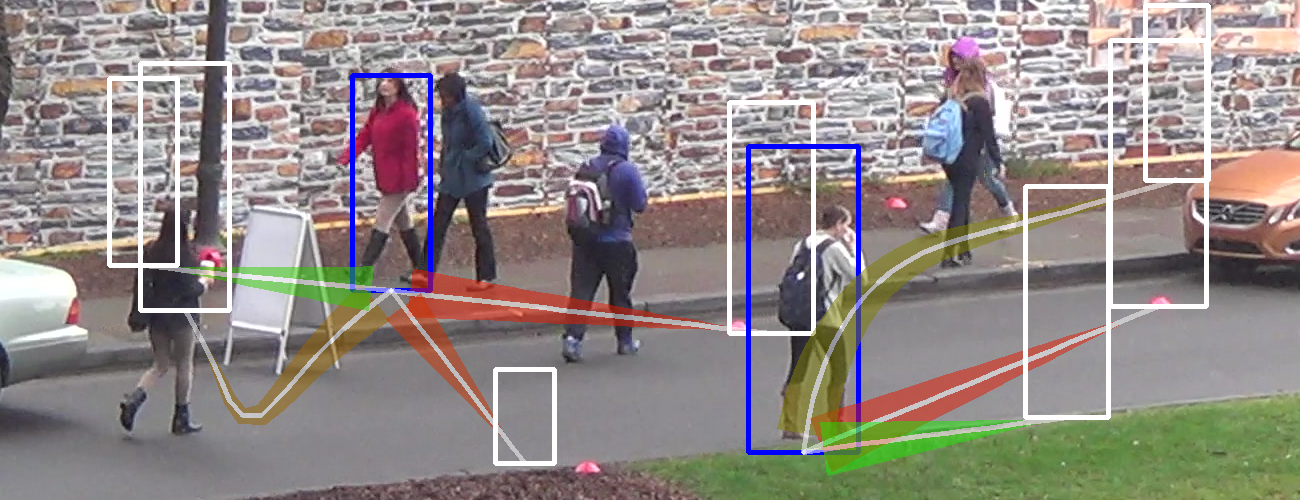} \\
    (a) & (b)\\
  \end{tabular}
  \vspace{-3mm}
  \caption{\small Network architecture and pruning mechanism. {\bf (a)} Tracklet features are passed through an embedding layer and then processed using a bi-directional LSTM. Its outputs are used to predict the IoU with ground truth bounding boxes ($iou$), presence of a person in a scene ($lab$), and regress bounding box shift to obtain ground truth bounding boxes ($sft$). {\bf (b)}  Candidate tracklets starting from two different bounding boxes in blue and ending with bounding boxes in white. In this case, during pruning phase the best tracklets, shown in green, are assigned the highest score and retained, and all others are eliminated. }
  \label{fig:network}
  \vspace{-4mm}
\end{figure*}


Ideally,  we should have $S(\Phi(\bT)) \approx {\mbox{\bf IDF}(\bT, \bG)}$ for every tracklet $\bT$ and the corresponding ground truth trajectory $\bG$. Unfortunately, at  inference time, $\bG$ is unknown by definition. To overcome this difficulty, recall from~\cite{Ristani16}  that ${\bf IDF}$ for tracklet $\bT=\left[\bd_1,\ldots,\bd_n\right]$ and ground truth trajectory $\bG=\left[\bg_1,\ldots,\bg_n\right]$ is defined as twice the number of detections matched by ground truth, over sum of total lengths of the two:
{\small
\begin{equation}
\mbox{\IDF}(\bT,\bG) = {2 \cdot \sum\limits_{n:\bd_n \not = \vv{0}, \bg_n \not = \vv{0}} \mathbbm{1}(IoU(\bd_n, \bg_n) > 0.5) \over |\{n : \bd_n \not = \vv{0}\}| + |\{n : \bg_n \not = \vv{0}\}|} \; ,
\label{eq:idf}
\end{equation}
}
where $IoU$ is the intersection over union of the bounding boxes. To approximate it without knowing $\bG$, we write
{\small
\begin{equation}
  S(\Phi(\bT)) = {2 \cdot \sum\limits_{n:\bd_n \not = \vv{0}, lab_n > 0.5} iou_n \over |\{n : \bd_n \not = \vv{0}\}| + |\{n : lab_n > 0.5\}|} \; ,
  \label{eq:appox}
\end{equation}
}
assuming that the deep network of Fig.~\ref{fig:network}, (a) has been trained to regress from $\bT$ to
\begin{itemize}

  \item $iou_n$: the prediction of intersection over union of the $\bd_n$ and $\bg_n$ boxes;

  \item $lab_n$: the prediction of whether the ground truth trajectory exists in frame $n$.

\end{itemize}
We  also train  our network  to predict  the necessary  change to  bounding box
$\bd_n$ to  produce the  ground truth  bounding box  $\bg_n$,  which we
denote  $sft_n$ .  It  is not  used  to compute  $S$, but  can  be used  during
inference to better align the observed bounding boxes with the ground truth.

To train the network to predict the $lab_n$, $iou_n$, and $sft_n$ values introduced above,  we define a loss function that is the sum of errors between predictions and ground truth. We write it as
{\small
\begin{align}
L(\mathcal{T} ,\mathcal{G}) &=
\sum_{n=1}^N L_{lab}(\bd_n, \bg_n) +
\sum\limits_{n : \bd_n \not = \vv{0}} L_{iou}(\bd_n, \bg_n)   \nonumber  \\
& + \sum\limits_{n : \bd_n \not = \vv{0}} L_{sft}(\bd_n, \bg_n), \label{eq:predLoss}\\
L_{lab}(\bd_n, \bg_n) & =  ||lab_n - \mathbbm{1}(\bg_n \not = \vv{0})||_2, \nonumber \\
L_{iou}(\bd_n, \bg_n) & =  ||iou_n - IoU(\bd_n, \bg_n)||_2,\nonumber \\
L_{sft}(\bd_n, \bg_n) & =  1 - IoU(\bd_n + sft_n, \bg_n), \nonumber
\end{align}
}
where $\bd_n+sft_n$  denotes the shifting  the bounding box  $\bd_n$ by  $sft_n$.

Arguably, we could have trained the network to directly regress to \IDF{} instead of first estimating $iou_n$,  $lab_n$,  and  $sft_n$ and then using the approximation of Eq.~\ref{eq:appox} to compute it. However, our experiments have shown that asking more detailed feedback for every time step, as we do, forces the network to better understand motion, while a good estimation of \IDF{} can be often produced by an average prediction.

We  chose  not  to apply  any  weight  factors to  the
components  of the  loss  function  because its  components  could  be seen  as
identifying the false  positive (when $lab$ should be zero)  and false negative
(when $IoU < 0.5$) errors, and since we wanted to weigh the two equally, we did
not use any weight factors to $L_{lab}$, $L_{sft}$, $L_{iou}$.

\subsection{Training Procedure}
\label{sec:training}

The key to avoiding exposure bias while training the network is to supply a rich training set. To this end, we alternate between the following two steps:
\begin{enumerate}

\vspace{-0.1cm}
  \item  Run the hypothesis generation algorithm of Section~\ref{sec:tracklets} using current network weights when evaluating $S$;

\vspace{-0.2cm}
  \item Add the newly created tracklets to the training set and perform a single epoch of training.
\vspace{-0.1cm}

\end{enumerate}
In addition  to learning the network  weights, this procedure helps  refine the
final  tracking result:  The tracking  procedure of  Section~\ref{sec:tracking}
makes discrete  choices about  which hypotheses  to pick  or discard,  which is
non-differentiable. We nevertheless  help it make the best  choices by training
the model  on all candidates, both  good and bad, encountered  during tracking.
In  other  words, our  approach  makes  discrete  choices during  training  and
then  updates the  parameters based  on all  hypotheses that  {\it could}  have
been  selected,  which  is  similar  in  spirit  to  using  a  straight-through
estimator~\cite{Bengio13b}.

While  simple  in   principle,  this  training  procedure   must  be  carefully
designed  for optimal  performance. We  list  here the  most important  details
of  our  implementation  and  study  their impact  in  the  ablation  study.

\parag{Stopping criterion.}  We start the  process with random  network weights
and stop  it when  the training  set size  increases by  less than  $5\%$ after
iterating the process  $10$ times. We then  fully train the model  on the whole
resulting training  set. This process can  be understood as a  slow traverse of
the  search space.  It  starts  with an  untrained  model  that selects  random
hypotheses. Then, as the training progresses, new hypotheses are added and help
the network both to differentiate between good and bad alternatives and to pick
the best ones with increasing confidence.

\parag{Randomized merging.} During inference, we  grow each tracklet by merging
it  with one  that  yields  the highest  possible  score.  By contrast,  during
training, we  make the  training set  more diverse  by randomizing  the merging
process.  To  do  that,  we  assign to  candidates  for  merger  a  probability
proportional  to softmax  of the  score of  the merged  result multiplied  by a
weight coefficient. We  initially set the coefficient so that  the optimal pair
is almost  always chosen and  we then progressively  reduce it to  increase the
randomness.


\parag{Balancing the dataset.} One potential  difficulty is that this procedure
may result in an unbalanced training set in terms of the \IDF{} values to which
we want to  regress. We solve this  by splitting the dataset into  10 groups by
\IDF{}  value ($[0.0;  0.1), [0.1;  0.2), \cdots,  [0.9, 1.0]$),  selecting all
samples  from the  smallest group,  and then  the same  number from  each other
group. This  enables us to  perform $h$-hard-mining  by selecting $h*K$  samples at
random and retaining the $K$ that contribute most to the loss.

\vspace{-0.2cm}

\section{Results}
\label{sec:results}

\vspace{-1mm}
We now present the datasets we use, baselines we compare against, our results, and finally a qualitative analysis.

\subsection{Datasets}
\label{subsec:datasets}

We used the following publicly available datasets to benchmark our approach.

\parag{\Duke{}}~\cite{Ristani16}. It contains 8 sequences,  with 50  minutes of  training data,  and testing sequences of 10 ("Hard", dense crowd traversing several camera views) and 25 minutes ("Easy") with hidden ground truth, at 60fps.

\parag{\MotSeven{}}~\cite{Milan16b}. It contains 7 training-testing sequence pairs with similar statistics and hidden ground truth for the test sequences, spanning 785 trajectories and both static and moving cameras. For each, there are 3 different sets of detections using different algorithms, which makes it possible to evaluate the quality of the tracking without overfitting to a specific detector.

\parag{\MotFive{}}~\cite{Leal-Taixe15}. It contains  11 training and 11  testing sequences, with moving and  stationary cameras  in various settings.  The ground truth  for testing is  hidden, and for  each testing sequence  there is a  sequence with roughly similar statistics in the training data.

\subsection{Baselines}
\label{sec:baselines}

We compared against a number of recent algorithms that collectively represent the state-of-the-art. We distinguish below  between those that do not use appearance cues and those that do.

\parag{Algorithms that ignore Appearance Cues.}
\vspace{-2mm}
\begin{itemize}[leftmargin=*]

 \vspace{-1mm}
 \item  \lp{}~\cite{Leal-Taixe15} is the highest-scoring appearance-less original baseline presented with MOT15. It formulates tracking in terms of solving a linear program.

\vspace{-1mm}
 \item \rnn{}~\cite{Milan17} relies on a recurrent neural network and is similar to ours in spirit because it uses RNN for tracking in a straightforward way. However it is trained using a different loss and approach to create the training data.

 \vspace{-1mm}
 \item \ptrack{}~\cite{Maksai17} aims to improve results of other methods by refining the trajectories they produce, to maximize an approximation of  the   \IDF{} metric. The approximation is hand-designed, and not learned as in our approach.

 \vspace{-1mm}
  \item \sort{}~\cite{Bewley16,Murray17} combines Kalman filtering with a Hungarian algorithm and currently is the fastest one on the \MotFive{} dataset.

\end{itemize}
\vspace{-2mm}

\parag{Algorithms that exploit Appearance Cues.}
\begin{itemize}[leftmargin=*]

  \vspace{-1mm}
  \item \mht{}~\cite{Yoon18} performs multiple hypothesis tracking, aided, among
others, by pose features extracted from convolution pose machines~\cite{Wei16}.

  \vspace{-1mm}
  \item  \cdsc{}~\cite{Tesfaye17} uses  domination  set  clustering to  perform
within- and  across-camera  tracking. It  employs  image features  from
ResNet-50~\cite{He16} pre-trained on ImageNet.

 \vspace{-1mm}
  \item \reid{}~\cite{Zhang17}  performs hierarchical clustering  of tracklets,
and  leverages  the re-identification  model of~\cite{Zhang17a}  pre-trained  on  7
different datasets.

 \vspace{-1mm}
  \item \bipcc{}~\cite{Ristani16}  clusters detections with  similar appearance
by solving a binary integer problem. This  is a baseline method for the \Duke{}
dataset.

 \vspace{-1mm}
  \item  \dman{}~\cite{Zhu18}  uses dual  attention  networks  to perform  data
association by focusing on relevant image parts and temporal fragments.

 \vspace{-1mm}  \item \jcc{}~\cite{Keuper18}  handles multiple  object tracking
 and motion  segmentation as  a joint  co-clustering problem.  It solves  it by
 local search to group pixels and  bounding boxes. This returns both tracks and
 segmentation.

 \vspace{-1mm}   \item   \motdt{}~\cite{Long18}  performs   hierarchical   data
association by  grouping detections  using a learned  re-identification metric,
exploiting geometric features, and Kalman filter.

 \vspace{-1mm}
 \item \mhtblstm{}~\cite{Kim18}  resembles our  approach in  spirit. It  uses a
multiple hypothesis tracker and a sequence  model to score the tracks. However,
it is trained using only ground-truth  sequence with at most one false positive
and sometimes missed detections.

 \vspace{-1mm}
 \item \edmt{}~\cite{Chen17c}  relies on a multiple hypothesis   tracker.  Its growing  and   pruning  phase depend on  learned detection-detection  and detection-scene  association  models that are used to  better score detections and hypotheses.

 \vspace{-1mm}
 \item \fwt{}~\cite{Henschel18}  solves a binary quadratic problem to
optimally group head and body detections, obtained separately.
\end{itemize}
We will  show in Section~\ref{sec:results}  that we outperform  both classes of methods when using the same setting as they do.

\subsection{Experimental Protocol}
\label{sec:protocol}

In this section, we describe the features we use in practice along with our approach to batch processing, training, and choosing hyperparameters.

\parag{Features}

For a fair comparison against the two classes of baselines described above, we use either features in which appearance plays no part or features that encode actual image information. We describe them below.

\subparagraph{Appearance-less features.}
\vspace{-2mm}

We  use the following simple  features  that  can be  computed  from  the detections  without further  reference  to the  images:
\begin{itemize}[leftmargin=*]
\item Bounding  box coordinates  and  confidence ($\in  \mathbbm{R}^5$).
\item Bounding  box  shift  with  respect  to  previous  and next  detection in  the  tracklet  ($\in \mathbbm{R}^8$).
\item Social  feature - a description  of the detections in  the vicinity, $\in
\mathbbm{R}^{3*M}$.  It comprises  offsets to  the $M$  nearest detections  and
their confidence values. All  values are expressed relative to image
size for better generalization.
\end{itemize}

\subparagraph{Appearance-based features.}
\vspace{-2mm}

As a basis  for appearance, we used the 128-dimentional  vector produced from a
bounding box  by the  re-identification model of~\cite{Hermans17}.  Distance in
euclidian  space  between  such  vectors  indicate  similarity  between  people
appearances  and likelihood  that they  are the  same person.  To this  end, we
provide following additional features in our appearance-based model:
\begin{itemize}[leftmargin=*]
\item Appearance vector for each bounding box ($\in \mathbbm{R}^{128}$).
\item Euclidian distance from appearance in  the bounding box to the appearance
that best  represents trajectory  so far  before the current  batch, if  one is
available ($\in  \mathbbm{R}^1$). To pick  the appearance that  best represents
trajectory  so  far, we  computed  euclidian  distances  between each  pair  of
appearances  in  the trajectory,  and  picked  one  with  the smallest  sum  of
distances to all others.
\item Crowd density feature - distance  from the center of current bounding box
to the  center of  nearest 1st, 5th,  and 20th detection  in the  current frame
($\in \mathbbm{R}^3$). As  we discuss in the ablation study,  that feature made
impact  on the  behavior  of our  model  with appearance  in  very dense  crowd
scenarios.
\vspace{-2mm}
\end{itemize}

\parag{Batch processing.}

In Section~\ref{sec:tracking}, we focused on  processing a batch of $N$ images.
In  practice, we  process longer  sequence by  splitting them  into overlapping
batches, shifting  each one by  ${N \over  3}$ frames. While pruning hypotheses, we never  suppress all those that can be merged with trajectories from the previous batch.
This ensures that we can incorporate all  tracks from  the previous  batch. We  used 3-second  long
batches for  training as  in~\cite{Sadeghian17}. During inference,  we observed
that our model is  able to generalize beyond 3s, and  having longer batches can
be  beneficial  in cases  of  long  occlusions.  Inference used  6-second  long
batches.

\parag{Training and Hyperparameters}

For  all  datasets  and  sequences, cross-validation revealed that  thresholds  $C_{IoU}$  and
$C_{score}$ of Sec.~\ref{sec:tracklets} equal to  $0.6$ and the hard-mining parameter
$h$ of Sec.~\ref{sec:training} equal to 3 to be near-optimal choices. For  {\bf
DukeMTMC},  we selected  a validation  set of  15'000 frames  for each  camera,
pre-trained the model on data from  all cameras simultaneously, and performed a
final training on the training data  for each individual sequence. We used
only \Duke{} training data to train the appearance model of~\cite{Hermans17}.  For each \MotFive{} pair of training and testing sequence pair, we used the training sequence for validation purposes and the remaining training sequences to learn the network weights.  For \MotSeven{}, we pre-trained our model on \PathTrack{}, the appearance model of~\cite{Hermans17} on on {\bf CUHK03}~\cite{Li14d} dataset,  and used the \MotSeven{} training sequences for validation purposes. More details are in the appendix.

\subsection{Comparative Performance}
\label{sec:perf}

 \begin{table}
\vspace{-4mm}
 \begin{center}
  \setlength\tabcolsep{2.1pt}
  \begin{tabular}{|l|c|c|c|c|c|c|c|}    
  \hline
  Method & App. &  \IDF & \MOTA & IDs  & \IDF & \MOTA & IDs \\ \hline
  Sequence & & \multicolumn{3}{c|}{Easy} & \multicolumn{3}{c|}{Hard} \\ \hline

 \ours{}       & + &\bf 84.0 & \bf 79.2 & \bf 169 & \bf 76.8 & \bf 65.4 & \bf 267  \\ \hline
 \mht{}        & + &80.3     & 78.3     & 406     & 63.5     & 59.6     & 1468     \\ \hline
 \reid{}       & + &79.2     & 68.8     & 449     & 71.6     & 60.9     & 572      \\ \hline
 \cdsc{}       & + &77.0     & 70.9     & 693     & 65.5     & 59.6     & 1637     \\ \hline
 \ours{}-geom  & - &76.5     & 69.3     & 426     & 65.5     & 59.1     & 972      \\ \hline
 \ptrack{}     & - &71.2     & 59.3     & 290     & 65.0     & 54.4     & 661      \\ \hline
 \bipcc{}      & + &70.1     & 59.4     & 300     & 64.5     & 54.6     & 652      \\ \hline

\end{tabular}
\caption{Results  on the \Duke{} dataset.  The second column indicates
whether or not the method uses appearance information.}
\label{tab:duke}
\end{center}
\vspace{-2mm}
\end{table}


\begin{table}
\vspace{-4mm}
  \setlength\tabcolsep{5.3pt}
\begin{tabular}{|l|c|c|c|c|}
  \hline
  Method  & Use appearance & \IDF     & \MOTA    & IDs   \\ \hline
  \ours{}-geom & - &\bf 27.1 & \bf 22.2 &  \bf 700  \\ \hline
  \sort{}      & - &26.8     & 21.7     &  1231 \\ \hline
  \rnn{}       & - &17.1     & 19.0     &  1490 \\ \hline
  \lp{}        & - &----     & 19.8     &  1649 \\ \hline
\end{tabular}
\caption{Results on the \MotFive{} dataset. Appearance is never used. }
\label{tab:mot15small}
\vspace{-2mm}
\end{table}


\begin{table}
\vspace{-2mm}
  \setlength\tabcolsep{4.8pt}
  \begin{tabular}{|l|c|c|c|c|}    
  \hline
  Method & Use appearance & \IDF & \MOTA & IDs \\ \hline

 \ours{}        & + &\bf 57.2 & 44.2 & \bf 1529 \\ \hline
 \dman{}        & + &55.7     & 48.2     & 2194     \\ \hline
 \jcc{}         & + &54.5     & 51.2     & 1802     \\ \hline
 \motdt{}       & + &52.7     & 50.9     & 2474     \\ \hline
 \mhtblstm{}    & + &51.9     & 47.5     & 2069     \\ \hline
 \edmt{}        & + &51.3     & 50.0     & 2264     \\ \hline
 \fwt{}         & + &47.6     & \bf 51.3 & 2648     \\ \hline

\end{tabular}
\caption{Results on the \MotSeven{} dataset. Appearance always used. }
\label{tab:mot17small}
\vspace{-4mm}
\end{table}

We compared  on \Duke{} and  \MotFive{} against methods that  ignore appearance
features because their results are reported on these two datasets. For the same
reason, we used  \Duke{} and \MotSeven{} to compare against  those that exploit
appearance.  We  summarize the  results  below,  reporting \IDF{}  and  \MOTA{}
tracking metrics, and  a number of identity switches (IDs),  and provide a much
more detailed breakdown in the appendix. We present some tracking
results in Fig.~\ref{fig:resultsDuke}  and Fig.~\ref{fig:resultsMOT}.
\parag{Comparing to Algorithms that exploit Appearance.}

We report our results on \MotSeven{} in Tab.~\ref{tab:mot17small} and on  \Duke{} in Tab.~\ref{tab:duke}.

On \Duke{},  our approach performs  best both for  the Easy and  Hard sequences
in  terms  of  \IDF{},  \MOTA{},  and the  raw  number  of  identity  switches.
Furthermore,  unlike  other  top  scoring methods  that  use  re-identification
networks pre-trained  on additional datasets,  ours was trained using  only the
\Duke{} training data.

On \MotSeven{}, our  approach is best both  in terms of \IDF{}  metric, and the
number of  identity switches. However,  it does poorly on  \MOTA{}. Strikingly,
\fwt{} does the exact opposite: it yields  best \MOTA{} and the worst \IDF{} on
this dataset.  Careful examination  of the trajectories  shows that  this comes
from  producing many  short trajectories  that increase  the overall  number of
tracked  detections, and  therefore  \MOTA{},  at the  cost  of assigning  many
spurious  identities, increasing  fragmentation,  and  decreasing \IDF{}.  This
example illustrates why we believe \IDF{}  to be the more meaningful metric and
why we have designed our tracklet scoring function to be a proxy for it.

\parag{Comparing to Algorithms that ignore Appearance.}
We report our results on \MotFive{} in~Tab.~\ref{tab:mot15small} and on \Duke{}
in Tab.~\ref{tab:duke}. On  \MotFive{} dataset, method most similar  to ours is
\rnn{}, which  also uses an RNN  to perform data association.  Despite the fact
that \rnn{} uses  external data to pre-train  their model, and we  use only the
\MotFive{}  training  data, our  approach  is  able  to  outperform it  with  a
large margin.  Another interesting comparison  is with \sort{},  which performs
nearly as  good as  our approach.  However, it can  not leverage  training data
effectively,  and  to show  that  we  additionally  ran  this approach  on  the
validation data  we used for  \Duke{}, where there  is much more  training data
that in \MotFive{}. This resulted in a  \MOTA{} score of 49.9 and \IDF{} one of
24.9, whereas our method reaches 70.0 and 74.6 on the same data.

Remarkably,  on \Duke{}  dataset, even  though  we ignored  appearance for  the
purpose  of this  comparison,  our  approach also  outperforms  or rivals  some
the  methods that  exploit it~\cite{Ristani16,Tesfaye17}.  This shows  that our
training procedure is powerful enough to overcome this serious handicap.


\begin{figure}
  \includegraphics[width=\columnwidth,trim={12cm 15cm 8cm 10cm},clip]{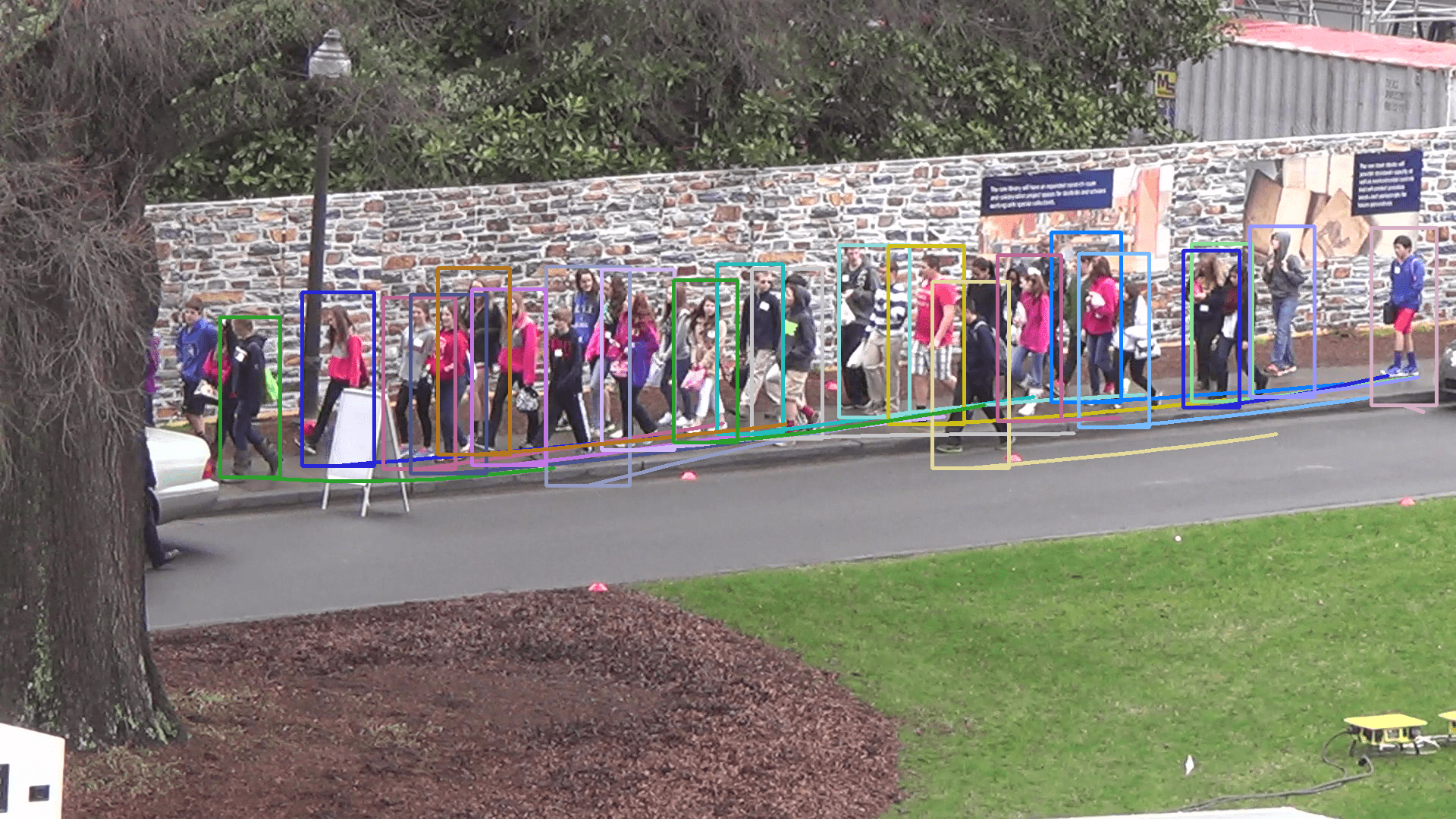}
  \caption{\small Bounding boxes and the last 6 seconds of tracking, denoted by lines, in dense crowd on \Duke{} dataset.}
  \label{fig:resultsDuke}
\end{figure}

\begin{figure}
\begin{tabular}{c @{\hspace{0.5\tabcolsep}} c}
  \hspace{-3mm}
  \includegraphics[width=0.495\columnwidth,trim={0 0 0 0},clip]{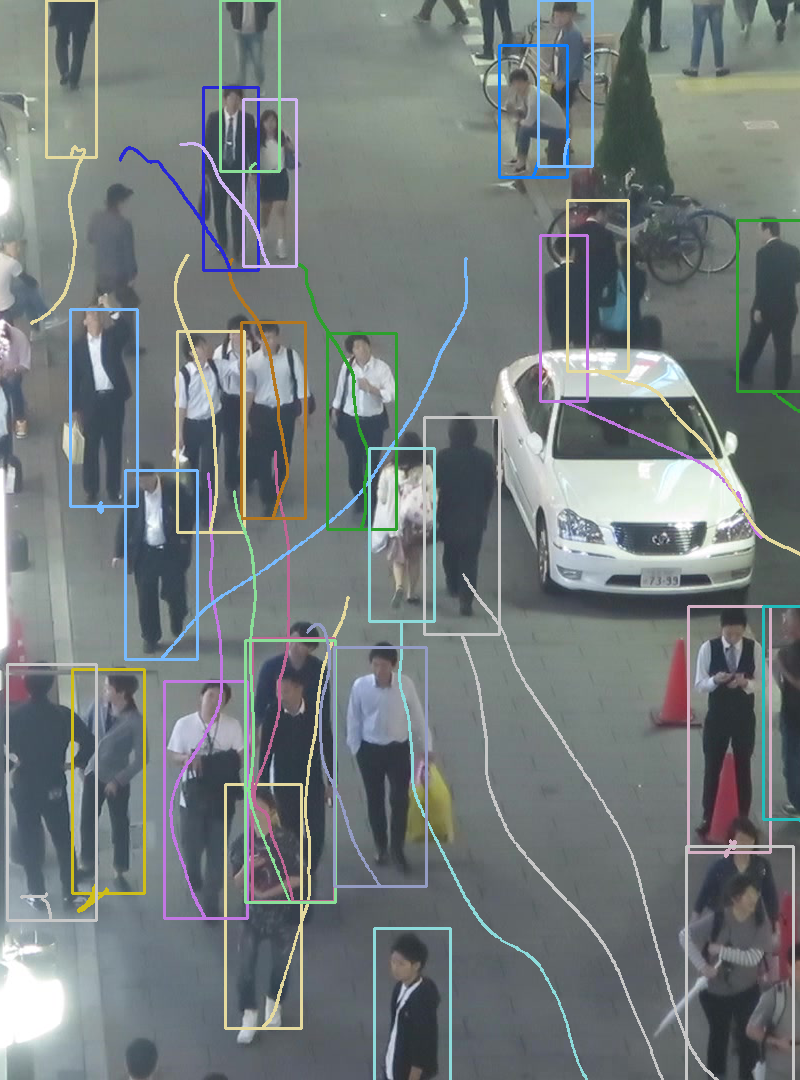} &
  \includegraphics[width=0.495\columnwidth,trim={39.5cm 0 0 0},clip]{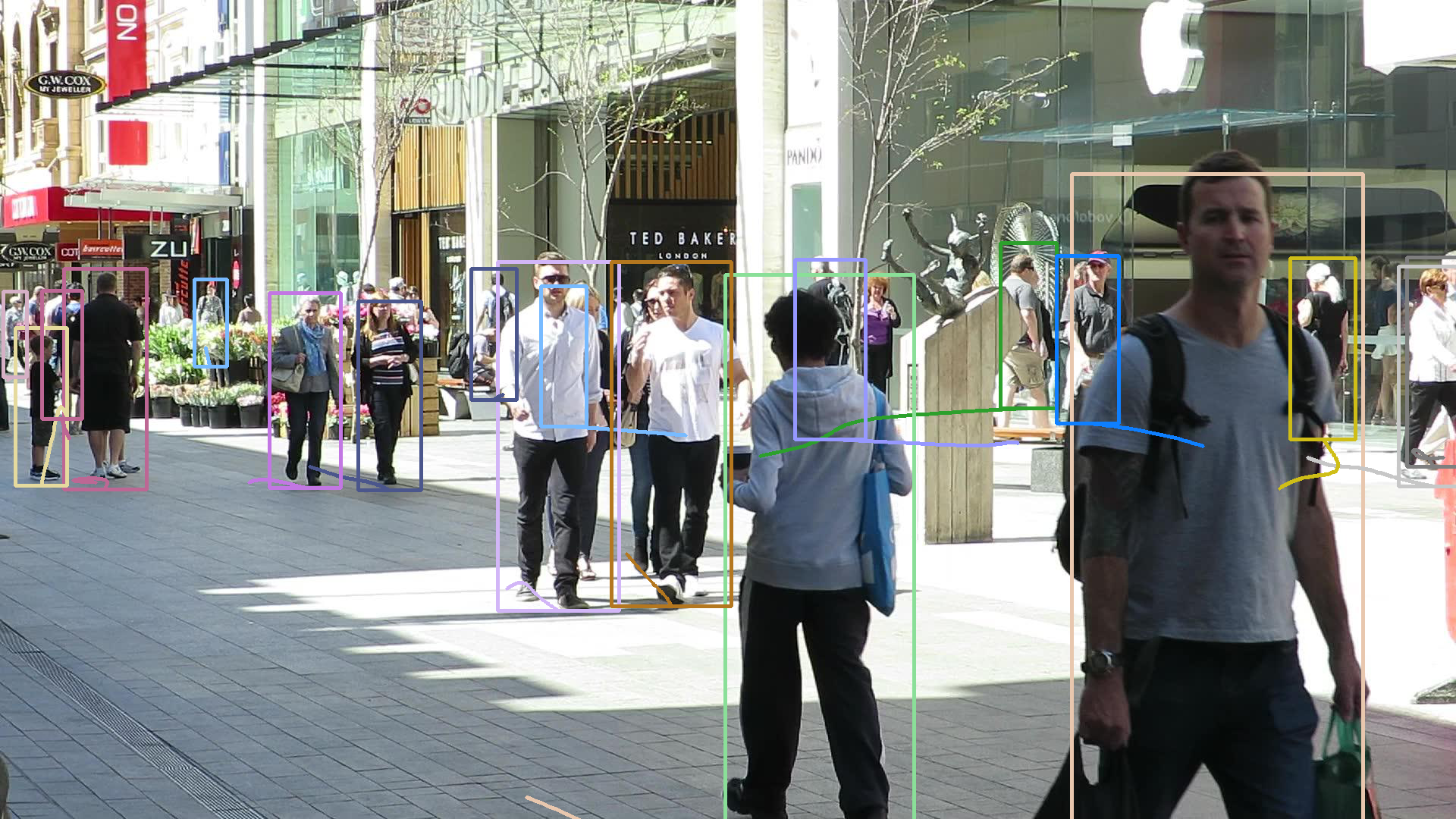} \\
\end{tabular}
  \caption{Bounding boxes and last 6 seconds of tracking, denoted by lines, in two sequences of the \MotSeven{} dataset.}
  \label{fig:resultsMOT}
  \vspace{-5mm}
\end{figure}

\subsection{Analysis}

We now analyze briefly some key components of our approach and provide additional details in the appendix.

\parag{Computational Complexity.}

We performed training  on a single 2.5Hz CPU, and  all other actions (computing
\IDF{}  values  for  dataset  balancing, generating  training  data,  etc.)  in
parallel on  20 such CPUs.  Training data contained  at most $1.5  \times 10^7$
tracklets (\Duke{} dataset, camera 6), resulting  in at most $1.35 \times 10^6$
training data points after balancing the dataset. Generating training data took
under 6  hours, and training on  it achieved best validation  scores within 30
epochs, taking  under 10  minutes each.  Inference runs at  about 2  frames per
second. However,  adding a  cutoff on  sequence scores in  the pruning  step of
Sec.~\ref{sec:candidates} speeds up our python  implementation to 30fps, at the
cost of a very small performance decrease (\IDF{} of 71 instead of 74.6).

\parag{Ablation study.}

The  last  15'000  frames  of  training sequences  of  \Duke{}  were  used  for
an  ablation  study. We  varied  the  three  main  components of  our  solution
to  show their  effect  on  the tracking  accuracy:  data composition,  scoring
function, and  training procedure. We report  the drop in \IDF{}  when applying
such  changes.  Creating   a  fixed  training  set   by  considering  tracklets
with  at  most  one  identity switch  as  in~\cite{Sadeghian17,Ma18}  decreased
performance (-3.9).  Pruning hypotheses  based on their  scores or  total count
like~\cite{Yoon18}  resulted in  either  a computational  explosion or  reduced
performance (-20). Computing loss on the prediction of $S(\Phi(T))$, regressing
\IDF{} value directly, not regressing bounding  box shifts, or using a standard
classification  loss as  in~\cite{Sadeghian17} were  equally counter-productive
(-5.1,  -13.2, -2.2,  -32.8).  Not  balancing the  training  set  or not  using
hard-mining also  adversely affected  the results  (-4.7, -2.5).  Selecting the
final  solution  using  an  Integer  Program instead  of  a  greedy  algorithm,
pre-training model with each type of  features separately, or training a deeper
network had no significant effect.

\parag{Feature  groups.}  We also  performed  an  evaluation of  how  different
features affect the quality of the solution. {\it Appearance features} improved
overall \IDF{} from  74.6 to 82.5, with appearance distance  feature having the
biggest effect. Crowd density feature  mostly affected crowded scenarios, where
our merging procedure  preferred to merge detections that are  further apart in
time, but more  visually similar, compared to less crowded  scenarios, where it
preferred  to  merge  detections  based  more on  the  spatial  vicinity.  {\it
Social  feature} mostly  affected  appearance-less model,  helping to  preserve
identities by ensuring that detections of the surrounding people are consistent
throughout trajectory, improving  \IDF{} from 67.5 to  74.6. {\it Probabilistic
merging} from  Sec.~\ref{sec:training} was vital to  fuse appearance-based and
geometry-based features together.  Without it, picking only  the best candidate
resulted in a model that performed merges mostly either based on the appearance
information (largely  ignoring spatial vicinity),  or based on the  spatial and
motion information (largely ignoring appearance information).

\vspace{-0.2cm}

\section{Conclusion}
\label{sec:conc}

We  have  introduced  a  training   procedure  that  significantly  boosts  the
performance of sequence models by iteratively  building a rich training set. We
have also  developed a sophisticated model  that can regress from  tracklets to
the \IDF{}  multiple target tracking  metric. We  have shown that  our approach
outperforms  state-of-the-art ones  on several  challenging benchmarks  both in
scenarios where appearance is used and where  it is not. In the second case, we
can even  come close to what  appearance-based method can do  without using it.
This could prove extremely useful to solve problems in which appearance is hard
to use, such as cell or animal tracking~\cite{Milan17}.

 In future work,  we will extend our data association  procedure to account for
more advanced appearance  features, such as 2D  and 3D pose. We  will also look
into further reducing the loss-evaluation  mismatch by using the actual \IDF{},
instead  of our  proposed  \IDF{} regressor,  which would  require  the use  of
reinforcement learning.

\appendix
\section{An appendix}

\subsection{Ablation study}
\label{sec:ablation}

\begin{table*}
\begin{center}
\begin{tabular}{|l|l|c||l|c||l|c||l|c|}
  \hline
  \bf \# & \bf $\Delta$ Dataset & \bf  IDF & \bf $\Delta$ Loss & \bf  IDF & \bf$\Delta$ Training & \bf  IDF & \bf $\Delta$ Tracking & \bf IDF \\ \hline
  1 & Dataset: all pairs   & 71.5             & Loss on IDF      & 69.5                  & -hardmining           & 72.1 & Batch 6 & 72.6 \\
  2 & Dataset: mix of two  & 70.7             & Regressing IDF   & 63.4                  & -balanced dataset      & 69.9 & IP solution & 73.8 \\
  3 & Selected only        & 63.6             & -bbox regression & 72.4                  & pretraining          & 71.9 & & \\
  4 & Prunning by score    & ----             & -bbox loss       & 66.3                  & 2 layer LSTM & 74.1  & & \\
  5 & Prunning by count    & 54.2             & classification   & 41.8                  & & & & \\ \hline
\end{tabular}
\caption[Ablation study:  variations of training procedure,  loss, and training
data]{Ablation study. Left,  middle and right columns show  possible changes in
dataset creation procedure, loss function,  training and tracking procedure, as
well  as respective  values  of  {\bf IDF}  metric  with  respect to  reference
solution  ({\bf   IDF}  74.6).   Details  about  each   change  are   given  in
Sec.~\ref{sec:ablation}.}
\label{tab:ablation}
\end{center}
\end{table*}

\paragraph{Ablation  Study.}
It was  performed  on the  validation  data  of  \Duke{} using the last 15000 training frames in each camera view.

Tab.~\ref{tab:ablation} depicts the results organized in four columns.
\begin{itemize}

  \item {\bf Changes in dataset.} We quantify the impact of degrading the training dataset generation procedure of Section 3, by:
  \begin{enumerate}
    \item using random tracklets between all pairs of detections;
    \item using tracklets obtained by combining at most two ground truth trajectories;
    \item adding to the training data not all tracklets observed during growing phase, but only those present in the final solution;
    \item doing prunning using predicted score of the tracklet as cutoff;
    \item doing prunning by retaining fixed number of tracklets with best scores;
  \end{enumerate}

  1), 2), and  3) yield a smaller  and less diverse training data,  which had a
detrimental effect on the results of tracking. 4) did not allow us to train any
reasonable model,  because of the  computational explosion of  the trajectories
with very  similar scores, that  were all taken  into training data.  5) proved
ineffective for  the same reason  - training  data contained many  very similar
trajectories.

\item {\bf Changes in the loss function.} We modify the loss function, described in Section 4.1 by:
  \begin{enumerate}
    \item using loss of $||IDF(D, T) - S(\Phi(D))||_2$;
    \item regressing the value  of $IDF$ directly,  without splitting  the task into  accounting for false  positives or  false negatives;
    \item not modifying  the  input detections based on the  regression of bounding box shifts;
    \item removing $L_{sft}$ component from the  loss function;
    \item posing task  as a classification task,  where tracklet belongs to the positive class {\it iff} all detections overlap with some ground truth trajectory with $IoU$ of at least  0.5.
  \end{enumerate}
      1) resulted  in small decrease,  probably due  to the fact  that multiple
loss  components acted  as regularizers.  2) gave  even worse  results, because
understanding the behaviour of $IDF$ function is much harder than understanding
behaviour  of false  positives and  false negatives,  which we  regress through
$lab$  and  $iou$. Difference  between  3)  and  4)  shows that  simply  having
$L_{sft}$  as part  of the  loss  function improves  the results,  acting as  a
regularizer.  5) doesn't  result  in a  very  good trained  model  due to  many
overlapping  sequences, some  of which  have $IoU$  greater than  0.5 in  every
frame, and some don't, and it is  hard for the model to distinguish between the
two.

\item {\bf Changes in the training procedure.} We modify the training procedure of Section 4.2 by:

  \begin{enumerate}
    \item not using hard-mining;
    \item not balancing the dataset;
    \item pre-training embedding for appearance and geometric features separately;
    \item using 2 layer LSTM, instead of a single layer, as depicted in Fig.2, (a);
  \end{enumerate}

\item {\bf Changes in the tracking procedure.} We modify the tracking procedure of Section 3 by:
  \begin{enumerate}
    \item using shorter batch in tracking (3s, same as during training, instead of 6s);
    \item selecting final solution by an IP trying to maximize objective of Eq.~1, rather than adding trajectories one by one greedily;
  \end{enumerate}

\end{itemize}

Additionally, while  it may seem  logical to use  $L_{iou}$ to predict  the IoU
between the modified  bounding box $\bd_n+sft_t$ and the  ground truth bounding
box $\bg_n$, in practice that makes it  harder to train the network as if finds
an easy solution of regressing empty bounding boxes, which never intersect with
the  ground  truth, thus  always  making  a  perfect prediction  of  $L_{iou}$.
Instead,  we use  the  network during  inference in  the  autocontext mode:  we
predict  the bounding  boxes, update  the input  tracklet with  them, and  then
regress the intersection over union of the new tracklet to compute the value of
$S$.

\subsection{Detailed Benchmark Results}
\label{sec:benchmark}


\begin{table*}[!b]
  \begin{center}
{\small
\begin{tabular}{|l|l|c|p{9.2cm}|}
  \hline
  Measure &	Better &	Perfect &	Description \\ \hline
  MOTA &	higher &	100 &	Multiple Object Tracking Accuracy~\cite{Bernardin08}. This measure combines three error sources: false positives, missed targets and identity switches. \\ \hline
  MOTP &	higher & 	100 &	Multiple Object Tracking Precision~\cite{Bernardin08}. The misalignment between the annotated and the predicted bounding boxes. \\ \hline
  IDF1&	higher &	100 &	IDF~\cite{Ristani16}. The ratio of correctly identified detections over the average number of ground-truth and computed detections. \\ \hline
  FAF &	lower	& 0 &	The average number of false alarms per frame.\\ \hline
  MT &	higher &	100 & Mostly tracked targets. The ratio of ground-truth trajectories that are covered by a track hypothesis for at least 80\% of their respective life span.\\ \hline
  ML &	lower	& 0 &	Mostly lost targets. The ratio of ground-truth trajectories that are covered by a track hypothesis for at most 20\% of their respective life span.\\ \hline
  FP	& lower &	0 &	The total number of false positives. \\ \hline
  FN &	lower &	0 &	The total number of false negatives (missed targets). \\ \hline
  ID Sw. &	lower &	0 &	The total number of identity switches. \\ \hline
  Frag. &	lower &	0 &	The total number of times a trajectory is fragmented (i.e. interrupted during tracking).\\ \hline
  Hz &	higher &	Inf. &	Processing speed (in frames per second excluding the detector) on the benchmark. \\ \hline
\end{tabular}
}
\caption{Metrics description.}
\label{tab:legend}
\end{center}
\end{table*}

\begin{table*}[!b]
\begin{center}
{\small
\begin{tabular}{|l|c|c|c|c|c|c|c|c|c|c|c|c|}
\hline
Method & IDF1 &	IDP &	IDR &	MOTA &	MOTP &	FAF &	MT &	ML &	FP &	FN &	ID Sw. &	Frag. \\ \hline
\ours{} & 84.0 & 89.4 & 79.2 & 76.0 & 76.0 & 0.09 & 950 & 72 & 66,783 & 186,974 & 169 & 1256    \\ \hline
\mht{} & 80.3 & 87.3 & 74.4 & 78.3 & 78.4 & 0.05 & 914 & 72 & 35,580 & 193,253 & 406 & 1,116    \\ \hline
\reid{} & 79.2 & 89.9 & 70.7 & 68.8 & 77.9 & 0.07 & 726 & 143 & 52,408 & 277,762 & 449 & 1,060  \\ \hline
\cdsc{} & 77.0 & 87.6 & 68.6 & 70.9 & 75.8 & 0.05 & 740 & 110 & 38,655 & 268,398 & 693 & 4,717\\ \hline
\ours{}-geom & 76.5 & 83.9 & 70.3 & 69.3 & 74.8 & 0.10 & 813 & 89 & 76,059 & 248,224 & 426 & 2,081       \\ \hline
\ptrack{} & 71.2 & 84.8 & 61.4 & 59.3 & 78.7 & 0.09 & 666 & 234 & 68,634 & 361,589 & 290 & 783\\ \hline
\bipcc{} & 70.1 & 83.6 & 60.4 & 59.4 & 78.7 & 0.09 & 665 & 234 & 68,147 & 361,672 & 300 & 801 \\ \hline
\end{tabular}
}
\caption{Full benchmark results on Easy set of sequences of {\bf DukeMTMC} dataset.}
\label{tab:duke-easy}
\end{center}
\end{table*}

\begin{table*}[!b]
\begin{center}
{\small
\begin{tabular}{|l|c|c|c|c|c|c|c|c|c|c|c|c|}
\hline
Method & IDF1 &	IDP &	IDR &	MOTA &	MOTP &	FAF &	MT &	ML &	FP &	FN &	ID Sw. &	Frag. \\ \hline
\ours{} & 76.8 & 89.3 & 67.4 & 65.4 & 75.3 & 0.12 & 450 & 87 & 35,596 & 210,639 & 267 & 977    \\ \hline
\reid{} & 71.6  &  85.3 & 61.7 & 60.9 & 76.8 & 0.14 & 375 & 104 & 40,732 & 237,974 & 572 & 993    \\ \hline
\ours{}-geom & 65.5 & 79.3 & 55.8 & 59.1 & 74.0 & 0.14 & 379 & 102 & 39,576 & 251,256 & 972 & 1,855    \\ \hline
\cdsc{} & 65.5 & 81.4 & 54.7 & 59.6 & 75.4 & 0.09 & 348 & 99 & 26,643 & 260,073 & 1,637 & 5,024   \\ \hline
\ptrack{} & 65.0 & 81.8 & 54.0 & 54.4 & 77.1 & 0.14 & 335 & 104 & 40,978 & 283,704 & 661 & 1,054  \\ \hline
\bipcc{}  & 64.5 & 81.2 & 53.5 & 54.6 & 77.1 & 0.14 & 338 & 103 & 39,599 & 283,376 & 652 & 1,073\\ \hline
\mht{}  & 63.5 & 73.9 & 55.6 & 59.6 & 76.7 & 0.19 & 400 & 80 & 55,038 & 231,527 & 1,468 & 1,801 \\ \hline
\end{tabular}
}
\caption{Full benchmark results on Hard set of sequences of {\bf DukeMTMC} dataset.}
\label{tab:duke-hard}
\end{center}
\end{table*}

\begin{table*}[!b]
\begin{center}
{\small
\begin{tabular}{|l|c|c|c|c|c|c|c|c|c|c|c|}
\hline
Method & MOTA & IDF1	 &MT	 &ML &	FP&	FN &	ID Sw. &	Frag. &	Hz & Hardware \\ \hline
\ours{}-geom & 22.2 & 27.2 & 3.1 & 61.6 & 5,591 & 41,531 & 700  & 1,240  & 8.9 & 2.5 GHz CPU \\ \hline
\sort{} & 21.7 & 26.8 & 3.7 & 49.1 & 8,422 & 38,454 & 1,231  & 2,005  & 1,112.1  &1.8 GHz CPU \\ \hline
\lp{}  & 19.8 & --- & 6.7 & 41.2 & 11,580 & 36,045 & 1,649  & 1,712  & 112.1 & 2.6Hz 16 CPU \\ \hline
\rnn{} & 19.0 & 17.1 & 5.5 & 45.6 & 11,578 & 36,706 & 1,490  & 2,081  & 165.2 & 3GHz, CPU \\ \hline
\end{tabular}
}
\caption{Full benchmark results on {\bf MOT15} dataset.}
\label{tab:mot15}
\end{center}
\end{table*}

\begin{table*}[!b]
\begin{center}
{\small
\begin{tabular}{|l|c|c|c|c|c|c|c|c|c|}
\hline
Method & Cam1 & Cam2 & Cam3 & Cam4 & Cam5 & Cam6 & Cam7 & Cam8 & Overall \\ \hline
\ours{} & 82.1/76.8 & 70.9/67.7 & 88.5/84.0 & 70.9/63.0 & 58.9/49.9 & 88.5/81.6 & 71.4/71.6 & 66.0/66.4 & 74.6/70.1 \\ \hline
\sort{} & 23.7/37.8 & 27.7/51.2 & 26.5/53.0 & 25.7/40.5 & 28.6/68.0 & 23.0/54.9 & 27.6/56.8 & 16.4/37.0 & 24.9/49.9 \\ \hline
\end{tabular}
}
\caption{Comparison to \sort{} method on the validation data for DukeMTMC dataset, {\bf IDF}/{\bf MOTA}.}
\label{tab:sort}
\end{center}
\end{table*}

\begin{table*}[!b]
\begin{center}
{\small
\begin{tabular}{|l|c|c|c|c|c|c|c|c|c|}
\hline
Method & MOTA & IDF1 & MT\% & ML\% & FP & FN & IDs & Frag & Hz \\ \hline
\ours{} &    44.2 &   57.2 &    16.1 &    44.3 &   29,473 &    283,611 &    1,529 &    2,644 &    4.8 \\ \hline
\dman{} &    48.2 &   55.7 &    19.3 &    38.3 &   26,218 &    263,608 &    2,194 &    5,378 &    0.3 \\ \hline
\jcc{} &    51.2 &   54.5 &    20.9 &    37.0 &   25,937 &    247,822 &    1,802 &    2,984 &    1.8 \\ \hline
\motdt{} &    50.9 &   52.7 &    17.5 &    35.7 &   24,069 &    250,768 &    2,474 &    5,317 &    18.3 \\ \hline
\mhtblstm{} &    47.5 &   51.9 &    18.2 &    41.7 &   25,981 &    268,042 &    2,069 &    3,124 &    1.9 \\ \hline
\edmt{} &    50.0 &   51.3 &    21.6 &    36.3 &   32,279 &    247,297 &    2,264 &    3,260 &    0.6 \\ \hline
\fwt{} &    51.3 &   47.6 &    21.4 &    35.2 &   24,101 &    247,921 &    2,648 &    4,279 &    0.2 \\ \hline
\end{tabular}
}
\caption{Full benchmark results on {\bf MOT17} dataset.}
\label{tab:mot17}
\end{center}
\end{table*}

Here  we   now give a  description of tracking metrics  in  Tab.~\ref{tab:legend} and  full
results for  all benchmarks in  Tab.~\ref{tab:duke-easy}, ~\ref{tab:duke-hard},
~\ref{tab:mot15}. Legend  information and  results for \MotFive{}  dataset were
collected from the benchmark website~\url{https://motchallenge.net/} on the 6th
of May, 2018, while results for \MotSeven{} and \Duke{} datasets were collected
on the 30th of October, 2018. Our tracker results are available there under the
names {\bf SAS} and {\bf SAS\_full} for \Duke{} benchmark, {\bf SAS\_MOT15} for
\MotFive{} benchmark, and {\bf SAS\_MOT17} for \MotSeven{} benchmark.

We also report results  of our comparison to \sort{} on  the validation data we
used for DukeMTMC dataset in Tab.~\ref{tab:sort} We tuned the parameters of the
method ({\it max\_age}, {\it min\_hits},  detection quality cutoff) on the same
data we used for training for ablation study, using grid search.

\subsection{Training Protocol}
\label{sec:details}

We have trained the model with Adam  with the fixed learning rate of 0.001. Our
embedding  layer consists  of  a fully  connected layer,  followed  by a  batch
normalization layer. Size of the hidden state of LSTM were 300. In all cases we
kept  $C_{iou}=0.6$, $C_{score}=0.6$  and  trained with  batches  of length  3s.
Thanks to abundance  of training data, we  used fps of 3  for {\bf DukeMTMC} dataset.
For \MotFive{}  and  \MotSeven{} datasets,  we
trained  the  model with  the  maximum  frequency  every sequence  allowed,  to
increase the amount of training data.
During inference, we used batches of length  6s. We used the bounding box shift
regression  only in  combination with  the DPM~\cite{Felzenszwalb08}  detector, as  for other
types of detectors it did not  prove useful. Nevertheless, we kept $L_{sft}$ as
a part of  a loss function. We  plan to make our implementation  (in Python and
using Tensorflow) publicly available upon acceptance of the paper.

For  each {\bf  MOT15} sequence  group (KITTI,  ADL, etc.),  we trained  on all
sequences  excluding the  group, using  them for  validation purposes,  and ran
inference on the test  sequences from the same group. For  {\bf MOT17}, we used
{\bf  PathTrack} for  pre-training of  the  model, and  training sequences  for
validation. We trained re-identification network on {\bf CUHK03} dataset.

The coefficient, which we used to  multiply the probabilities before softmax to
allows the probabilistic merging, described in  the paper, was annealed from 10
to 0.1 in 30 epochs.

{\small
\bibliographystyle{ieee}
\bibliography{string,vision}
}
\end{document}